\documentclass[letterpaper, 10 pt, conference]{ieeeconf}

\IEEEoverridecommandlockouts  
\usepackage{cite}
\usepackage{amsmath,amssymb,amsfonts}
\usepackage{algorithmic}
\usepackage{graphicx}
\usepackage{textcomp}
\usepackage{xcolor}
\usepackage{url}
\usepackage{footnote}
\usepackage{graphics}
\usepackage{subfigure}

\usepackage{multirow}
\usepackage{booktabs}
\usepackage{tabularx}

\usepackage[ruled,linesnumbered,vlined]{algorithm2e} 
\def\BibTeX{{\rm B\kern-.05em{\sc i\kern-.025em b}\kern-.08em
    T\kern-.1667em\lower.7ex\hbox{E}\kern-.125emX}}
\title{\LARGE \bf 
Active Loop Closure for OSM-guided Robotic Mapping \\ 
in Large-Scale Urban Environments
}

\author{
Wei Gao$^{1}$, Zezhou Sun$^{1}$, Mingle Zhao$^{1}$, Cheng-Zhong Xu$^{1}$, and Hui Kong$^{1*}$
\thanks{*This work was supported in part by the Science and Technology Development Fund of Macau SAR under Grants 0046/2021/AGJ/ and 0067/2023/AFJ.}
\thanks{*Corresponding author.}
\thanks{$^{1}$Wei Gao, Zezhou Sun, Mingle Zhao, Chengzhong Xu, and Hui Kong are with the State Key Laboratory of Internet of Things for Smart City (SKL-IOTSC), Faculty of Science and Technology, University of Macau, Macao, China. ({\tt\small \{gw.ga0.wei, zhao.mingle\}@connect.umac.mo, \{zzsun, czxu, huikong\}@um.edu.mo})}}
\begin{document}

\maketitle  
\thispagestyle{empty}
\pagestyle{empty}
\begin{abstract}

The autonomous mapping of large-scale urban scenes presents significant challenges for autonomous robots. To mitigate the challenges, global planning, such as utilizing prior GPS trajectories from OpenStreetMap (OSM), is often used to guide the autonomous navigation of robots for mapping. However, due to factors like complex terrain, unexpected body movement, and sensor noise, the uncertainty of the robot's pose estimates inevitably increases over time, ultimately leading to the failure of robotic mapping. To address this issue, we propose a novel active loop closure procedure, enabling the robot to actively re-plan the previously planned GPS trajectory. The method can guide the robot to re-visit the previous places where the loop-closure detection can be performed to trigger the back-end optimization, effectively reducing errors and uncertainties in pose estimation. The proposed active loop closure mechanism is implemented and embedded into a real-time OSM-guided robot mapping framework. Empirical results on several large-scale outdoor scenarios demonstrate its effectiveness and promising performance. 
\end{abstract}

\section{Introduction}

Autonomous mapping of unknown environments by autonomous robots is an important research field in robotics society owing to its potential applications in tasks such as search and rescue, map-based navigation, and intelligent construction. This mapping process is usually achieved through the exploration by a robot in a specified unknown environment. In exploration, robots typically perceive unknown environments through their onboard sensors and construct maps by self-planning algorithms and simultaneous localization and mapping (SLAM) technologies. Over decades, many representative exploration methods have been proposed\cite{yamauchi1997frontier,keidar2014efficient,cao2021tare,kulkarni2022autonomous}. However, autonomous exploration in large urban scenes remains challenging primarily due to the following issues. 

The first issue is about the efficiency of robotic mapping. 
As an autonomous mapping method, the robot is not allowed to continue to explore a limited region endlessly.
In general, exploration strategy can play a key role in exploration efficiency. In a large-scale unknown environment (even in urban communities), there may be many exploration routes (strategies), and the variance among them can be very large. Robots can choose meaningless paths by trying and erroring among different routes without global guidance. It usually takes forever and eventually leads to failure with the power exhaustion. So far, choosing a good exploration route (strategy) without global guidance is still an open issue in large-scale robotic exploration of unknown environments. 

Second, some existing exploration algorithms do not consider the uncertainty of robot pose estimation and the quality of maps \cite{keidar2014efficient,sun2020frontier,zhu2021dsvp, kulkarni2022autonomous, sun2023concave}. Generally, they assume that pose estimation is perfect and ready for the other exploration modules. However, the front-end pose estimation module often suffers from rhythmic shaking of the robot body, environmental degradation, sensor noise, and accidental movements when avoiding unexpected collisions, leading to large pose drift over time. 
Although a few methods start considering pose and map uncertainty during exploration \cite{soragna2019active,xu2021crmi,bai2023graph}, due to the lack of a priori global information, these methods can only balance the exploration efficiency and pose estimation accuracy within a short planning horizon. These methods do not consider it from a global perspective, resulting in the pose optimization being too localized after loop closure if any, and remaining a large global inconsistency in pose estimation. 

To deal with the above issues, we propose an OSM-guided framework, enabling the robot to accurately and efficiently map the environment along major roads in large-scale urban scenarios.
OpenStreetMap\footnote{OpenStreetMap, Available at https://www.openstreetmap.org} is a free geographic database and contains GPS data of the main roads in urban areas, providing global guidance for robotic mapping. Specifically, our contributions include:

\textbf{First}, to accommodate both mapping efficiency and accuracy, we propose to generate the shortest global trajectories in the OSM topological road networks based on Hierholzer’s algorithm \cite{jungnickel2005graphs}. It takes instant pose estimation accuracy into account by considerately incorporating local closed loops to avoid potential pose estimation crashes.  

\textbf{Second}, the GPS uncertainty in OSM can cause the following of the global trajectory to collapse. To deal with it, we propose an accurate and fast traversable-region segmentation scheme to optimize the local guidance point, with which the tracking of unreliable GPS trajectory is accompanied by the optimized guidance point within the traversable area. 

\textbf{Third}, due to the robot's body shaking, environmental degradation, sensor noise, and accidental movements when avoiding collisions, the uncertainty and error of the pose estimates inevitably increase over time. To address it, we propose a novel active loop closure procedure to re-plan the previously planned GPS trajectory timely based on an uncertainty evaluation metric, guiding the robot to travel to a previously visited location where loop-closure detection can take place to trigger the back-end optimization to reduce the error and uncertainty of pose estimation. 

\section{Related Work}

Mapping is an essential function of various types of mobile robots and is also important for navigation and localization. 
In recent years, utilizing autonomous robots for mapping unknown environments has made significant progress. Their application scenarios range from indoor scenarios \cite{umari2017autonomous, sun2020frontier} to real-world outdoor environments \cite{zhu2021dsvp, kulkarni2022autonomous, sun2023concave}. A common issue of these methods is that they seldom consider the pose estimation drift. In small scenes, the cumulative pose drift can be ignored because it is still acceptable for consistent mapping. However, when facing a large-scale scene, to avoid path-following failures caused by pose drift and to maintain the mapping quality, the robot needs to re-plan and control its motion to reduce pose uncertainty. This line of SLAM research is also referred to as active SLAM\cite{placed2023survey}. \cite{stachniss2005information} uses an efficient Rao-Blackwellized particle filter to represent the
posterior on maps and poses. It applies a decision-theoretic framework that simultaneously considers the uncertainty in the map and vehicle's pose to evaluate the potential
actions. \cite{zhang2022exploration} uses geometric observations for possible loop closure viewpoints actively in the exploration planning to improve the mapping result. \cite{tao20243d} exploits the semantic maps for active loop closure, allowing the robot to perform loop closure more effectively. 
However, due to the lack of a global view of the environment (i.e., without an initial global path planning), these methods perform active loop closure greedily at a local scale and might not obtain a globally consistent map. 

Alternatively, some other approaches consider more about autonomous robotic mapping from a global perspective with prior information (e.g., GPS information).  \cite{osswald2016speeding} directly exploits topology-metric graphs to optimize the robot exploration process. They used a frontier-based approach and solved the global planning with TSP to fully
explore the environment. However, pose uncertainty and map accuracy are not considered. Similarly, \cite{soragna2019active} aims to traverse all edges in the given topological-metric graph by solving the Chinese Postman Problem (CPP)\cite{eiselt1995arc}, and can actively trigger loop closure according to the pose uncertainty. However, this method lacks theoretical justification for establishing the connection between slam accuracy and the graph structure (proposed by \cite{placed2021fast}), which makes their method dependent on the choice of parameter($\lambda$) in their proposed core formulation when comparing candidate paths in some common cases.
Based on the theory proposed by  \cite{placed2021fast} and \cite{carrillo2012comparison},  \cite{bai2023graph} proposes a SLAM-aware planner that combines pose graph reliability with graph topology. This method can find an exploration path (by adding round-trip loop-closure paths on TSP routes) with the most informative and distance-efficient loop closures to globally enhance map quality. Although this method has been validated in simulation scenarios, the nodes of the prior topological graph in the simulated environment are very close to each other and the neighboring nodes can be reachable directly without considering traversability between nodes. In large-scale outdoor scenarios, the region between any two nodes might not be traversable, which means that we cannot greedily find surrounding nodes around the current node and add connections as edges. In addition, this method does not consider the uncertainty of the topological nodes. Therefore, this method may fail in large-scale outdoor scenarios and still needs validation in real-world scenes. 

In our case, we use OSM as a priori to provide a rough global topology of the scene, through which an initial global path can be planned before the robot starts the exploration. The potential problems such as the uncertainty in OSM data, the traversability analysis in complex terrain, and pose drift in a large-scale urban scene are all taken into consideration. Compared to the existing methods with active loop-closure modules\cite{zhang2022exploration,stachniss2005information,tao20243d}, our method can efficiently derive the routes that traverse all the roads and maintain a high-quality map with the help of the CPP algorithm and Theory of Optimal Experimental Design (TOED)\cite{pazman1986foundations} method at the global level. In the replanning stage whenever the robot undergoes a large uncertainty in pose estimation, our approach utilizes the Rural Postman Problem (RPP) to allow the robot to explore both the active loop closure and traverse the remaining roads efficiently. For real-world scenarios, we also propose a terrain analysis method and a planned-path tracking module to efficiently and accurately complete the map construction in real-world scenarios.

\begin{figure*} [t]
    \centering
    \vspace{-0.6cm}
    \includegraphics[width=0.83\linewidth]{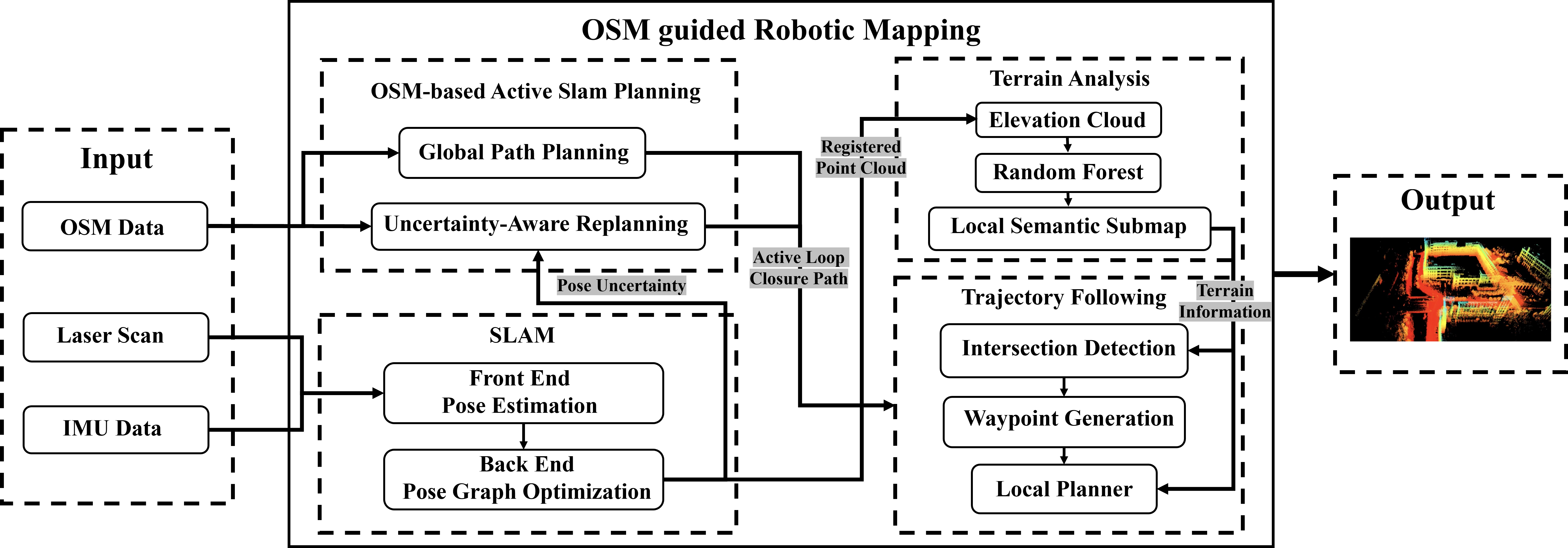}
    % \hspace{.2in}
    \caption{
    System overview of OSM-guided robot mapping framework.
   }
    \label{framework}
    \vspace{-0.5cm}
\end{figure*}

\section{Methodology}
\label{sec: Methodology}
\subsection{System Overview}
    Our overall framework is shown in Fig.\ref{framework}. Our system mainly consists of OSM-based active planning, SLAM, terrain analysis, and trajectory following modules. 
    The system initializes by completing the global path planning with OSM as a priori information. Once starting following the global path, the robot performs a real-time full SLAM Module (including front-end odometry, loop closure detection, and graph optimization) and returns the uncertainty of the current pose estimation. Whenever the uncertainty of the pose estimation reaches a certain level, the robot triggers the active loop closure procedure to regenerate the traversal order of the OSM road network, actively planning to go to an optimal previously visited location where a loop closure can occur. The pose drift error can be reduced promptly in this way. Then, from the loop-closure location, the robot restarts the traversal of the remaining road network, whose traversal order has also been updated in the active loop-closure procedure. With accurate pose estimation over time, the robot also maintains a local terrain semantic submap through random forests. Within the segmented traversable area, the robot detects the skeleton direction,  generates the optimal waypoint to go, and passes it to the local planner to complete the trajectory following. Our whole system runs in real time and can obtain an accurate point cloud map.

\subsection{OSM-based Active Loop Closure}

\subsubsection{Problem Formulation}

This part involves finding a closed path where the robot navigates along the edges of the OSM graph from the starting node, traverses each edge at least once, and goes back to the starting node, aiming at achieving high pose estimation accuracy throughout the process. Specifically, it can be described as follows:
\begin{itemize}
\item Global path planning: given a graph, aiming at finding a global path that traverses each edge at least once and achieving high instant pose estimation accuracy.
\item Uncertainty-Aware Replanning: given the global path, navigating with it. Whenever the robot's pose estimation uncertainty is high, finding an evolved optimal path that leads the robot to an optimal loop closure location quickly and allows for efficient traversal of unexplored edges.
\end{itemize}

For both problems mentioned above, we need an evaluation mechanism to evaluate the accuracy of SLAM estimates based on the graphical structure of SLAM problems constructed from different mapping routes.

Conventionally, this kind of mechanism can be formulated as the estimation-over-graphs (EoG) problem \cite{khosoussi2019reliable}. Next, we analyze the above two problems from this EoG perspective. Given the OSM graph $G$ and a candidate global route $\mathcal{R}$, when the robot navigates along with $\mathcal{R}$ and performs graph-based SLAM\cite{grisetti2010tutorial}, a pose graph can be incrementally built. We make an assumption similar to \cite{bai2023graph}, i.e., with the coarse structure of the prior OSM graph, we can obtain an approximate abstraction of the pose graph (called abstract graph) as a means of evaluating the reliability of the pose graph uncertainty. The formed abstract graph is denoted as $ G_{eog} = (V, E) $,
where $V$ is the set of $n$ nodes in the region to be explored and $E$ is the set of $m$ edges connecting the nodes. The travel distance of route $\mathcal{R}$ can be calculated by the function $Dis(R)$. We assume $G_{eog}$ is connected and the initial position of the robot is known.

The graph Laplacian matrix of $G_{eog}$ is defined as 
\begin{equation}
    \mathbf{L}_{o} \triangleq B_{o} B_{o}^{T} \in \mathbb{R}^{n \times n},
\end{equation}
where $B_{o} \in \mathbb{R}^{(n) \times m} $ is the incidence matrix of $G_{eog}$. 
The weighted Laplacian matrix of the graph $G_{eog}$ is defined as
\begin{equation}
    \mathbf{L}_{o} \triangleq B_{o} Diag \lbrace \omega_{1},\omega_{2}...\omega_{m} \rbrace B_{o}^{T} \triangleq \sum_{k=1}^{m} B_{k}B_{k}^{T} \cdot \omega_k ,
\end{equation}
where $\omega_{k}$ is the weight of edge $e_k$ in $G_{eog}$, $B_{k}$ is the $k$-th column vector of $B_{o}$.

According to the TOED, we use $D\text{-}opt$ to quantify the uncertainty of state vector estimation in the SLAM system, and it is defined as
\begin{equation}        
    D\text{-}opt(\mathbf{\Sigma})=\det(\mathbf{\Sigma})^{\frac1\ell}=\exp\left(\frac1\ell\sum_{k=1}^{\ell}\log(\lambda_k)\right),
    \label{D-opt}
\end{equation}
where $\ell$ is the dimension of the state vector, $\lambda_k$ is an eigenvalue of matrix $\mathbf{\Sigma}$. The $\boldsymbol{\Sigma}\in\mathbb{R}^{\ell\times\ell}$ is the covariance matrix measuring the uncertainty of the state vector. Essentially, the $D\text{-}opt$ measures the volume of the covariance ellipsoid, and it can capture global uncertainty and holds the monotonicity property  \cite{carrillo2012comparison}, which is a friendly score metric to evaluate the uncertainty of the pose-graph SLAM.

Next, we need to make a connection between the $\mathbf{L}_{o}$ and $\mathbf{\Sigma}$ so that we can estimate the uncertainty of the pose graph SLAM based on the graph structure. According to \cite{sorenson1980parameter}, the covariance matrix of any unbiased estimator of $x$, such as $\hat{x}$, satisfies $\Sigma_{\hat{x}} \geq \mathbb{I}^{-1}(x)$, where $\mathbb{I} (x)$ is the Fisher Information Matrix (FIM). 
The covariance matrix of the maximum likelihood estimation (MLE) of each pose is usually approximated
by evaluating $\mathbb{I}^{-1}(x)$ at the ML estimate $\hat{x}$. Furthermore, \cite{placed2022general} establishes the relationship between the FIM of the SLAM system and the Laplacian matrix of the underlying pose graphs, defined as follows,

\begin{equation}
\begin{aligned}
    D\text{-}opt(\mathbb{I}(x)) 
    &\approx D\text{-}opt(\frac{1}{2}\sum_{k=1}^{m} B_{k}B_{k}^{T} \cdot D\text{-}opt(\hat{\Sigma}_{k}^{-1})) \\
    &= D\text{-}opt(\mathbf{L}_{\gamma})
\end{aligned}
\end{equation}
where $\hat{\Sigma}_{k}$ is the covariance matrix for the $k$-th measurement, 
$B_{k}B_{k}^{T}$ is the $k$-th laplacian factor of the pose graph, and $\mathbf{L}_{\gamma} \in \mathbb{R}^{(n-1) \times (n-1)}$ is the weighted reduced Laplacian matrix (after anchoring a node) of the pose graph, and weighted by $D\text{-}opt(\hat{\Sigma}_{k}^{-1})$ of each edge in the pose graph. Note that we can now evaluate the uncertainty of the pose-graph SLAM based on the $\mathbf{L}_{\gamma}$ of $G_{eog}$ given different routes $\mathcal{R}$ of $G$. Therefore, we reformulate both Global path planning and Uncertainty-Aware replanning problems as follows:
\begin{itemize}
\item Global path planning: given an OSM graph $G$, find a route $\mathcal{R}$ from candidate routes set $\mathbf{R}$ that satisfies the Eq.\ref{global eq}.

\item Uncertainty-Aware Replanning:  given the global path $\mathcal{R}$, the part already traversed by the robot is denoted by $\mathcal{R}_1$, the part required to reach the loop closure location is denoted by $\mathcal{R}_2$, and the part to cover the rest edges (starting from the active loop-closure location) is $\mathcal{R}_3$. We aim to determine $\mathcal{R}_2$ and $\mathcal{R}_3$ by maximizing the $D\text{-}opt(\mathbf{L}_\gamma)$. 
\end{itemize}
\begin{equation}
\label{global eq}
\begin{aligned}
& \operatorname*{max}_{\mathcal{R} \in \mathbf{R}} & \quad & \frac{D\text{-}opt(\mathbf{L}_\gamma(\mathcal{R}))}{Dis(\mathcal{R})} \\
& \text{s. t.} & & \mathcal{R} \text{ is a route that} \\
& & & \text{traverses each edge of}\ G \text{ at least once}.
\end{aligned}
\end{equation}

\begin{figure*}[ht]
    \vspace{-0.2cm}
    \centering
    \begin{minipage}[]{0.28\linewidth}
        \centering
    \includegraphics[width=1\linewidth]{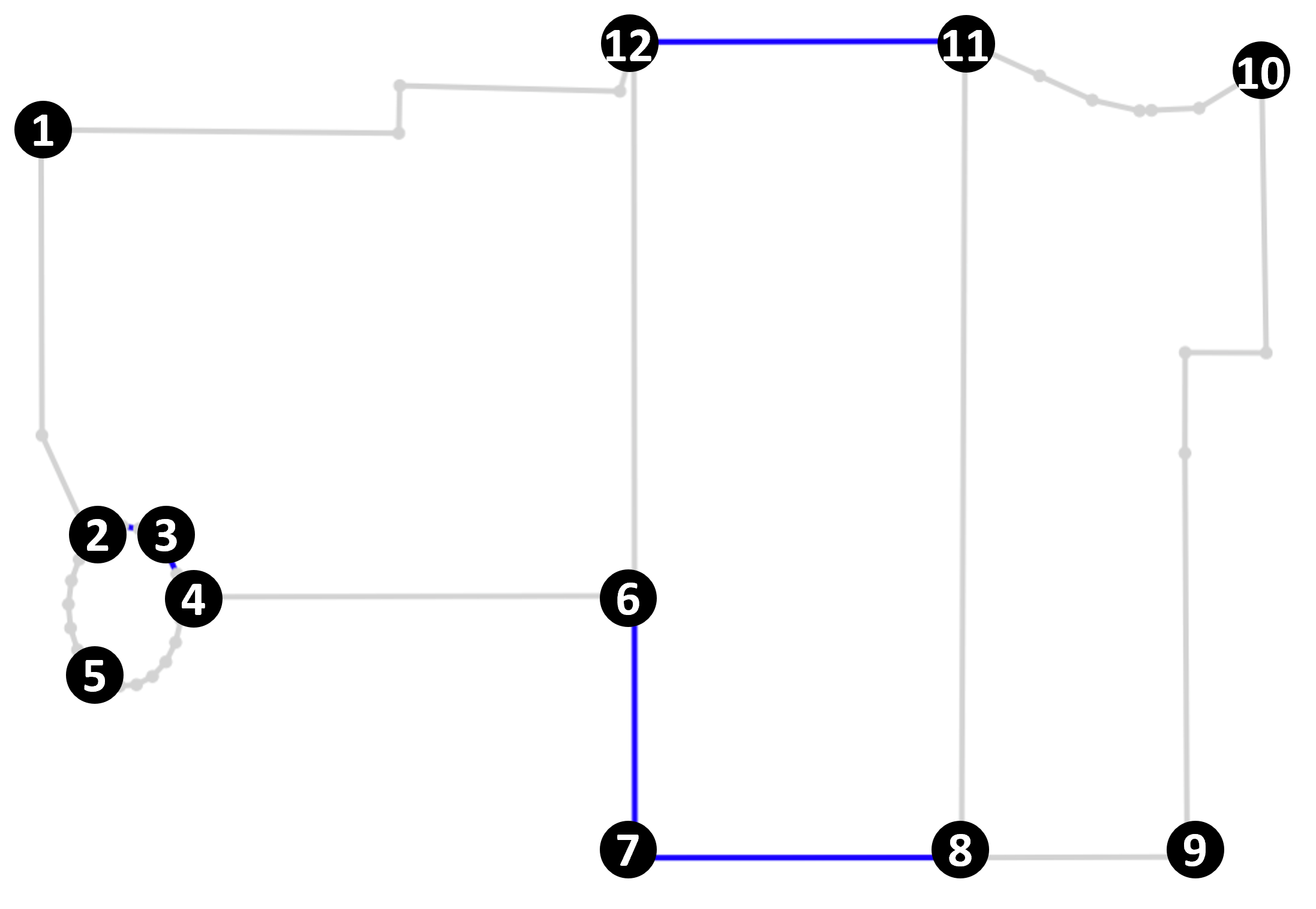}
    \end{minipage}
    \begin{minipage}[]{0.56\linewidth}
        \centering
        \subfigure[Routes with good loop closure constraints]{\includegraphics[width=1\linewidth]{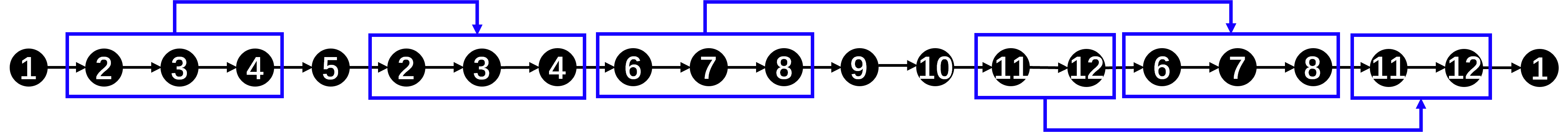}} 
        \subfigure[Routes with bad loop closure constraints]{\includegraphics[width=1\linewidth]{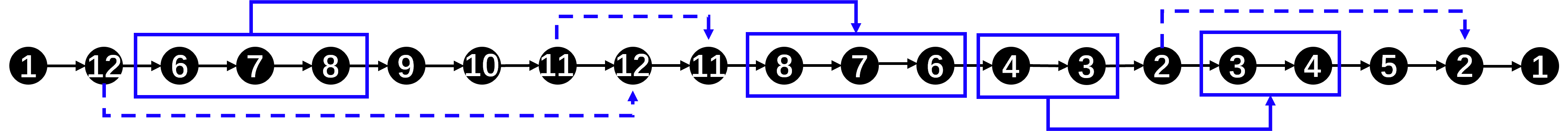}}
    \end{minipage}
    \caption{ 
    Illustration of two different graph structures constructed by two different global routes. The left graph corresponds to the OSM data. (a) and (b) are two routes obtained by the CPP solver as an example. 
    The blue solid lines represent the edges and the blue dashed lines represent the nodes where the robot has traversed duplicates, indicating potential loop closures at these locations.
    }
    \label{eog}
    \vspace{-0.5cm}
\end{figure*}

\subsubsection{Global Path planning}
\label{sec: Global Candidate Path Generation}
We formulate the global path traversal as CPP and use Hierholzer's algorithm\cite{jungnickel2005graphs} to find candidate routes based on three steps:
 1. Extract all the odd-degree nodes and connect these nodes to construct the complete graph. 
2. Compute the minimum weight match of the pairs of odd-degree nodes in the complete graph.
3. After augmenting the graph with matched pairs, we can identify a finite number of Eulerian circuits, each representing a candidate route.

In the real world, since the majority of OSM graphs are non-Eulerian, it is necessary to traverse certain edges and nodes more than once. During the mapping process, the repeated edges and the nodes with degrees greater than 2 will be traversed more than once to ensure that loop closure occurs at these locations.
Generally, different traversal routes of the robots can have different numbers of loop closures even if the total distance traveled is the same. 
In addition, different routes construct different graph structures, and the pose-graph SLAM accuracy after loop closure is usually different for different graph structures based on the TOED.

At the global path planning stage, the robot has not started the traversal yet. We assume that the abstract graph generated from the OSM graph is representative. To generate different routes, specifically as shown in Fig.\ref{eog}, we add different loop closure constraints onto the OSM graph. Through the graph structure, we construct different abstract graphs for different routes and their corresponding Laplacian matrices $\mathbf{L}_\gamma(\mathcal{R})$. Note that the traveled distance in different routes is the same, and the abstract graph generated from each candidate route has an equal number of nodes, which ensures that the Laplacian matrix constructed from the graph has the same dimensions. Therefore, we can directly evaluate different candidate routes by computing $D\text{-}opt(\mathbf{L}_\gamma(\mathcal{R}))$ and then select the optimal global route.

\subsubsection{Uncertainty-Aware Replanning}
\label{sec: Uncertainty-Aware Replanning}

The active loop closure procedure is essentially a re-planning process. During this process, the robot does not simply follow the previously planned global route. It needs to actively and promptly re-plan its path to a previously visited location. This re-planning is triggered by a mechanism that monitors the marginal covariance estimate from the pose graph optimization module. If the estimate is high, indicating a large uncertainty in the robot's pose estimation, the system initiates an active loop closure procedure to improve localization accuracy.

Different from the global planning, in the re-planning scenarios, the robot may have to traverse additional paths when heading towards loop closure points and while exploring the remaining edges. This results in each re-planned path having a varying number of edges and nodes, as well as Laplacian matrices with different dimensions. Therefore, the method used in global planning does not yield comparable $D\text{-}opt$ values for each path in the active replanning. 

To address this, our scheme is that when a loop closure occurs, mainly the historical keyframe poses between the two loop closure frames will be optimized, and the weight of the edge between the historical pose will update accordingly. According to Alg.\ref{alg:graph_edge_counts_update}, we can obtain the specific optimization counts $\mathbf{C}_\mathcal{R}$ for each edge with different routes (Including traveled routes and future routes). Next for visited and unvisited edges, we can update the weights accordingly
\begin{equation}
\label{eq:update_weights}
 \Omega(C_\mathcal{R}, \omega_{k0}) = \\
\begin{cases}
    D\text{-}opt(\hat{\Sigma}_{k}^{-1})\cdot  \frac{(\log_2(1 + C_\mathcal{R}))^{1/\alpha}}{(\log_2(1 + C_\mathcal{R}-n))^{1/\alpha}} & \text{visited} \\
    \omega_{k0} \cdot (\log_2(1 + C_\mathcal{R}))^{1/\alpha} & \text{unvisited}
\end{cases}
\end{equation}
where $n$ represents the optimal counts for traveled routes.  $D\text{-}opt(\hat{\Sigma}_{k}^{-1})$ is the true weight for the visited edges.  $\omega_{k0}$ indicates the estimated initial weight for each unvisited edge, which is inversely proportional to the distance.
Consequently, by calculating the $D\text{-}opt$, we can estimate the mapping accuracy by taking different paths. Importantly, although the number of edges and their connectivity have not been altered, the dimensions of the Laplacian matrix remain the same, making the $D\text{-}opt$ a meaningful metric for comparison.

\begin{algorithm}[ht]
    \caption{Update Graph Edge Optimal Counts }
    \label{alg:graph_edge_counts_update}
    \KwIn{\\
    \quad $G=(V,E)$, initial edge count with $\mathbf{C_0} = \mathbf{0}$\\
    \quad $\mathcal{R}$: a sequence of nodes indicate the path
    }
    \KwOut{Updated graph $G$ with new edge counts $\mathbf{C}_\mathcal{R}$}
    $\mathbf{C}_\mathcal{R} \leftarrow \mathbf{C_0}$ \\
    \For{$i \leftarrow 1$ \KwTo $\text{length}(\mathcal{R}) - 1$}{
        $u, v \leftarrow \mathcal{R}[i], \mathcal{R}[i+1]$\\
        \If{$(u, v) \in E$}{
            $\mathbf{C}_\mathcal{R}(u, v) \leftarrow \mathbf{C}_\mathcal{R}(u, v) + 1$\\
        }
    }
    $visited \leftarrow \{\}$ \\
    \For{$i \leftarrow 1$ \KwTo $\text{length}(\mathcal{R})$}{
        $n \leftarrow \mathcal{R}[i]$\\
        \If{$n \in visited$}{
            \For{$j\leftarrow i-1$ \KwTo $1$ }{
                \If{$\mathcal{R}[j] == n$}{
                    \ForEach{$k \leftarrow j$ \KwTo $i-1$}{
                        $\mathbf{C}_\mathcal{R}(\mathcal{R}[k], \mathcal{R}[k+1]) \leftarrow \mathbf{C}_\mathcal{R}(\mathcal{R}[k], \mathcal{R}[k+1]) + 1$\\
                    }
                    break\\
                }
            }
        }
        \Else{
            $visited \leftarrow visited \cup \{n\}$\\
        }
    }
    \Return $G=(V,E)$, $\mathbf{C}_\mathcal{R}$
\end{algorithm}

We illustrate our method with a simple example, as shown in Fig.\ref{eog}: Assume that a robot follows the path 1\(\rightarrow\)2\(\rightarrow\)3\(\rightarrow\)4\(\rightarrow\)5\(\rightarrow\)2\(\rightarrow\)3\(\rightarrow\)4\(\rightarrow\)6\(\rightarrow\)7\(\rightarrow\)8\(\rightarrow\)11, and is currently at node 11. At this node, the robot triggers the active loop closure re-planning phase, where it can choose any nodes from its historical trajectory as the loop closure location.
According to Alg.\ref{alg:uncertainty_aware_replanning}, the robot will consider the case of traveling to nodes 1, 2, 3, 4, 5, 6, 7, and 8, respectively, as well as evaluate which node would be the best choice to revisit to perform loop closure.

Let us consider one iteration in the loop of Alg.\ref{alg:uncertainty_aware_replanning} and take traveling to node 7 for performing loop closure as an example.
The problem now becomes traveling from node 11 to node 7 and then completing the remaining path. First, the robot needs to find the shortest path to node 7, the robot has two options: either 11\(\rightarrow\)12\(\rightarrow\)6\(\rightarrow\)7 or 11\(\rightarrow\)8\(\rightarrow\)7. Note that 
the edges 11\(\rightarrow\)12\(\rightarrow\)6 are unexplored while the edge 6\(\rightarrow\)7 is explored in the first option. In contrast, all the edges in the second option are explored. 
Although both paths are equal in terms of distance to be traversed, considering exploration efficiency and the limitations of backtracking for global pose optimization, we set lower costs for unexplored edges. Based on the Dijkstra's algorithm. The robot will incur extra travel distance as a trade-off to prioritize reaching target nodes via unexplored edges(\textit{function DijskraBestPath}). Therefore, at this moment, the robot will prefer the path 11\(\rightarrow\)12\(\rightarrow\)6\(\rightarrow\)7 ($\mathcal{R}_2$). Next, assuming the robot has arrived at node 7, the unexplored edges in the graph include 8\(\rightarrow\)9, 9\(\rightarrow\)10, 10\(\rightarrow\)11, 12\(\rightarrow\)1. These edges form the required graph (\textit{function CreateRequiredGraph}), while the already traversed edges constitute the optional edges. This problem thus becomes a variant of the RPP. However, unlike the typical RPP, our start and end points are not the same (in this case, the start is node 7 and the end is node 1) and the required graph may not be connected. To address this issue, we first compute the shortest distances between disconnected subgraphs to transform the required graph into a connected graph. Then, we add a virtual edge with minimal weight between the start and end nodes to convert the RPP into a variant of the CPP. At this point, we do not need to find an Eulerian path. Instead, we use the solution for CPP to find an Eulerian circuit, and then remove the virtual edge. This allows us to obtain the result $\mathcal{R}_3$. By comparing the $\mathcal{R}_2$, and $\mathcal{R}_3$ generated from each visited node, we can determine the best location for loop closure through Eq.\ref{objectiveFunction}.
\begin{equation}
    \label{objectiveFunction}
    % objective function:
    \max_{\mathcal{R}_2,\mathcal{R}_3}\; \frac{D\text{-}opt\left(\sum_{k=1}^{m} B_{k}B_{k}^{T}\cdot \Omega(C_\mathcal{R},\omega_k)\right)}{Dis(\mathcal{R}_1)+\sum_{e_k \in \mathcal{R}_2 \cup \mathcal{R}_3}Dis(e_k)}
\end{equation}

\begin{algorithm}[ht]
\caption{Uncertainty-Aware Replanning}\label{alg:uncertainty_aware_replanning}
  \KwIn{
    $G=(V,E,\omega,\mathbf{C}_0)$, initial edge count $\mathbf{C}_0$\\
    $\mathcal{R}_1$: a sequence of nodes that has been traveled
  }
  \KwOut{$\mathcal{R}_{opt}$: the routes that the robot will travel}
  UpdateNodeVisited($G$, $\mathcal{R}_1$)\\
  $max\_\mathcal{J} \leftarrow -\infty$\\
  % $max\_path_2 \leftarrow \{\}$\\
  % $max\_path_3 \leftarrow \{\}$\\
  $max\_path_2, max\_path_3 \leftarrow \{\}$\\
  $CurrentNode \leftarrow$ getCurrentNodeId($G$, $\mathcal{R}_1$)\\

  \ForEach{node $\in G$.nodes}{
    \If{node has been visited}{
      $\mathcal{R}_2 \leftarrow$ DijkstraBestPath($CurrentNode$, node)\\
      \ForEach{node $\in \mathcal{R}_2$}{
        node.visited $\leftarrow$ true\\
      }
      $G_{req} \leftarrow$ CreateRequiredGraph($G$)\\
      CreateConnectGraph($G_{req}$, $G$)\\
      $\mathcal{R}_3 \leftarrow$ RPPSolver(node, $G_{req}$)\\
      $\mathcal{R}_{full} \leftarrow \mathcal{R}_1 + \mathcal{R}_2 + \mathcal{R}_3$\\
      UpdateGraphEdgeCounts($G$, $\mathcal{R}_{full}$)\\

      $\mathcal{J} \leftarrow$ ObjectiveFunction($G$, $\mathcal{R}_{full}$)\\
      \If{$\mathcal{J} > \text{max\_}\mathcal{J}$}{
        % $max\_D\text{-}opt \leftarrow D\text{-}opt$\\
        $max\_path_2 \leftarrow \mathcal{R}_2$\\
        $max\_path_3 \leftarrow \mathcal{R}_3$\\
      }
    }
  }

  \Return{ $max\_path_2$, $max\_path_3$}
\end{algorithm}

\subsection{Terrain Analysis}
Given the OSM reference path, the robot needs to find the traversable area to follow the path. For this purpose, we propose a fast and accurate terrain analysis module with a low-cost Lidar and limited on-board computational resources. Unlike the commonly used LiDAR (e.g., the Velodyne Lidar), our LiDAR sensor is the Livox mid-360 with each frame containing a more sparse point cloud. To deal with the challenge in terrain analysis using this LiDAR,  we first transform point clouds in local-coordinate frames to the global frame and fuse the multi-frame point clouds near the robot position to construct a dense sub-map. In this way, we can compensate for the sparsity of LiDAR data. The point cloud is then voxelized and each voxel is dynamically updated so that the voxel contains the most recent data for any region within the local area. Since we use a ground-based robot, we reduce data redundancy by clipping and downsampling the elevation-point-cloud sub-map, keeping only the points whose vertical coordinates are within a certain range.

Up to this point, we can obtain a point-cloud sub-map, however, it is still not enough to meet our needs because common urban scenes have terrains such as grass, dirt, waterways, shrubs, and other areas that we do not want the robot to traverse through. These terrains are not easily distinguished from the road area based on geometric information alone. In addition, considering the limited onboard computational resources and the real-time requirements, we train a random forest classifier to classify the terrain. This algorithm naturally allows parallel prediction, and we assign the prediction task to multiple CPU cores to improve prediction speed. In our scenario, we classify terrain areas into three categories: roads, impassable areas (grass, shrubs, dirt, steep slopes, waterways, etc.), and obstacles.

\begin{figure} [ht]
    \centering
    \includegraphics[width = \linewidth]{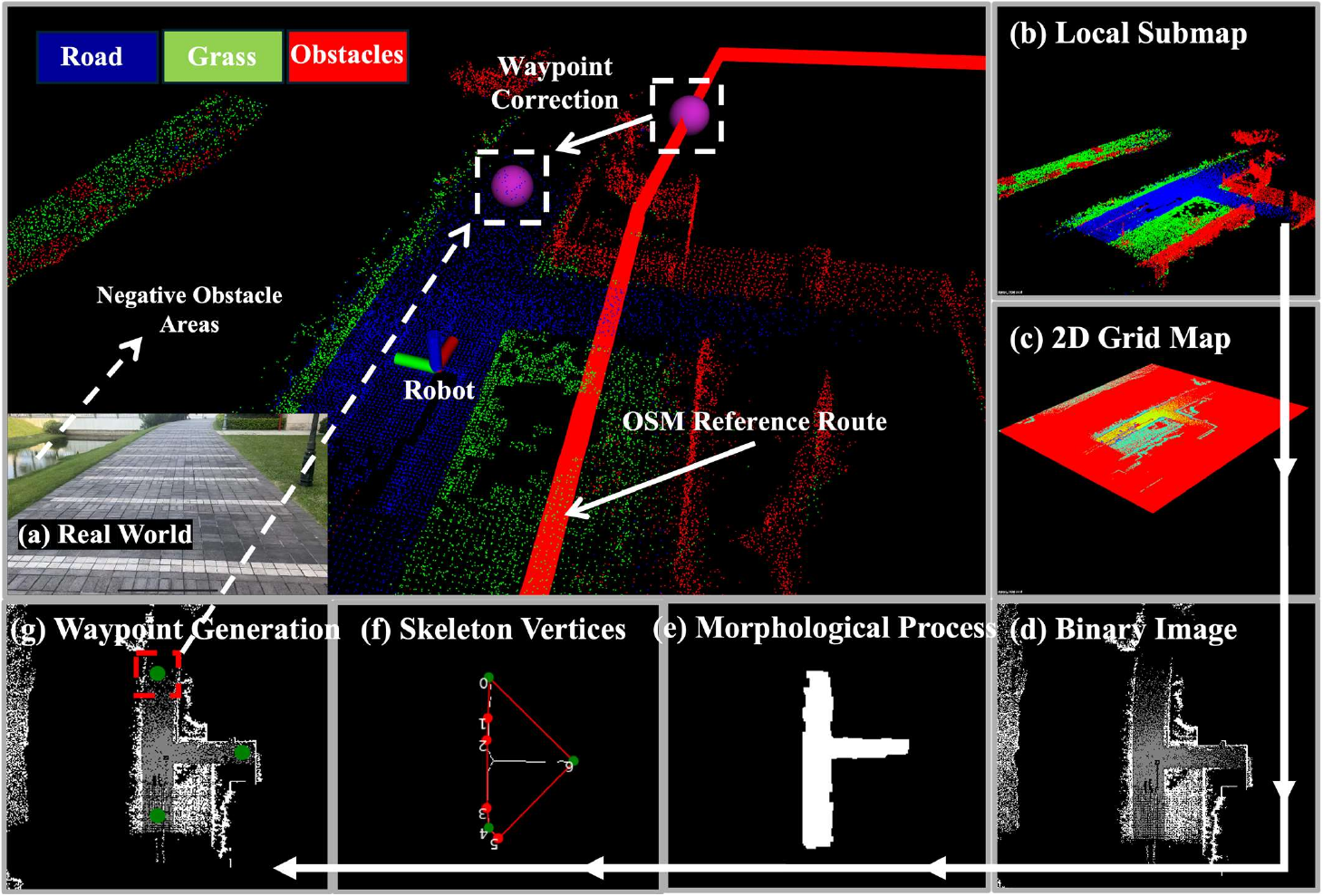}
   \caption{The process of waypoint generation. First, the semantic local sub-map is transformed into a 2D grid map after updating the grid state. Subsequently, the road area is extracted from the free-state regions of the grid map. Morphological operations are then applied to refine the road area. Finally, the skeleton is extracted and the convex hull vertices are calculated to generate waypoints within the current perception range.}
    \label{fig: Local Target Optimization}
    \vspace{-0.5cm}
\end{figure}

Specifically, we extracted a 4D feature vector for each point within each voxel: \textit{elevation}, the height difference of each point within a voxel relative to the lowest point in that voxel; \textit{planarity}, the smallest eigenvalue of the covariance matrix of all points' coordinates within a voxel, which describes the distribution of points inside; \textit{std\_dev}, the standard deviation of the z-coordinates within the voxel; \textit{intensity}, the reflectivity of the point. Subsequently, we used 30 frames of typical scenario data for training, with each point serving as a training sample, to obtain an ultra-lightweight terrain classification model. Our terrain classification module achieves about 50-FPS performance on an Intel i7 processor. 
Fig.\ref{fig: Local Target Optimization} shows a visualization of one example prediction result.

\subsection{Following OSM Reference Route}

Considering the existence of large uncertainty in OSM data and the deviation between OSM data and actual road topology, the waypoints provided by OSM mostly do not lie on road regions. For example, in \ref{fig: Local Target Optimization}, the purple dot lying on the OSM (red) route deviates far from the road (blue) region. Robot navigation can usually fail if only depending on these OSM guidance waypoints in the path following. Therefore, it is necessary to timely optimize the guidance waypoints based on the previously estimated terrain and the OSM waypoints such that the optimized waypoints lie in the traversable region and are consistent with the OSM waypoints in terms of guidance direction. For that, we extract all candidate intersections based on the estimated terrain type and derive the waypoints with the help of OSM information.

The overall process is shown in Fig. \ref{fig: Local Target Optimization}. First, we simplify the environment representation by converting the local semantic map into 2D grids based on the terrain type. Each grid is associated with a state: free, occupied, or unknown. The grid is dynamically updated based on the terrain analysis results. To obtain all intersection exits and road centers, we extract regions corresponding to free grids and perform morphological operations. The closing operation fills in the gaps caused by sparsity within the road regions. The open operation removes the noise and corrects some misclassified points. Then we extract the center point of the road-region's skeleton. Due to the irregularity, we do not assume that every endpoint of the skeleton branch is an intersection point. Therefore, we further extract the largest convex hull region from the skeleton branches, the vertex of this convex hull is viewed as the center of each intersection exits. We use a sliding time window to acquire the set of all intersection exits detected over a period, denoted as $\Theta$. After that, we assign a weight to each intersection to find out the next optimal waypoint $\mathbf{p}_{opt}$, which is calculated as 
\begin{equation}
\mathbf{p}_{opt} = \underset{\mathbf{\theta_i} \in \mathbf{\Theta}}{\arg\max} \left( K_{d} \cdot \frac{1}{1 + d_i} + K_{\alpha} \cdot \frac{1}{1 + \alpha_i} \right)
\end{equation} 
where $K_{d}$ and $K_{\alpha}$ are the weights to balance the importance of the distance and angle. $d_i$ denotes the distance from each intersection exit $\mathbf{\theta_i}$ to the last OSM points, and $\alpha_i$ represents the angle formed by the vector pointing from the robot to the intersection exit $\mathbf{\theta_i}$ and the vector pointing from the robot to the last OSM points.

\section{Experiments}

\subsection{Implementation}

Our approach is implemented and deployed on the quadruped robot platform. The devices include a Livox Mid-360 LiDAR, a built-in IMU, and an onboard computer with an Intel(R) Core(TM) i7-1165G7 CPU @ 4.70GHz running ROS.  
In the SLAM module, we adopt the pose estimation module by \cite{xu2022fast}, 
the LiDAR-Iris\cite{wang2020lidar}-based loop detection 
and GTSAM-based\cite{dellaert2012factor} pose-graph optimization. 
In the planning module, we efficiently implement the representation and the CPP and RPP solver. The system runs in real-time. 
In the terrain analysis and trajectory follow modules, based on \cite{cao2022autonomous}, we have made improvements so that the terrain analysis can include semantic information and the target waypoint can be calculated by combining the semantic information. 

\subsection{Real World Test}

\begin{figure}[ht]
    \centering
    \includegraphics[width = \linewidth]{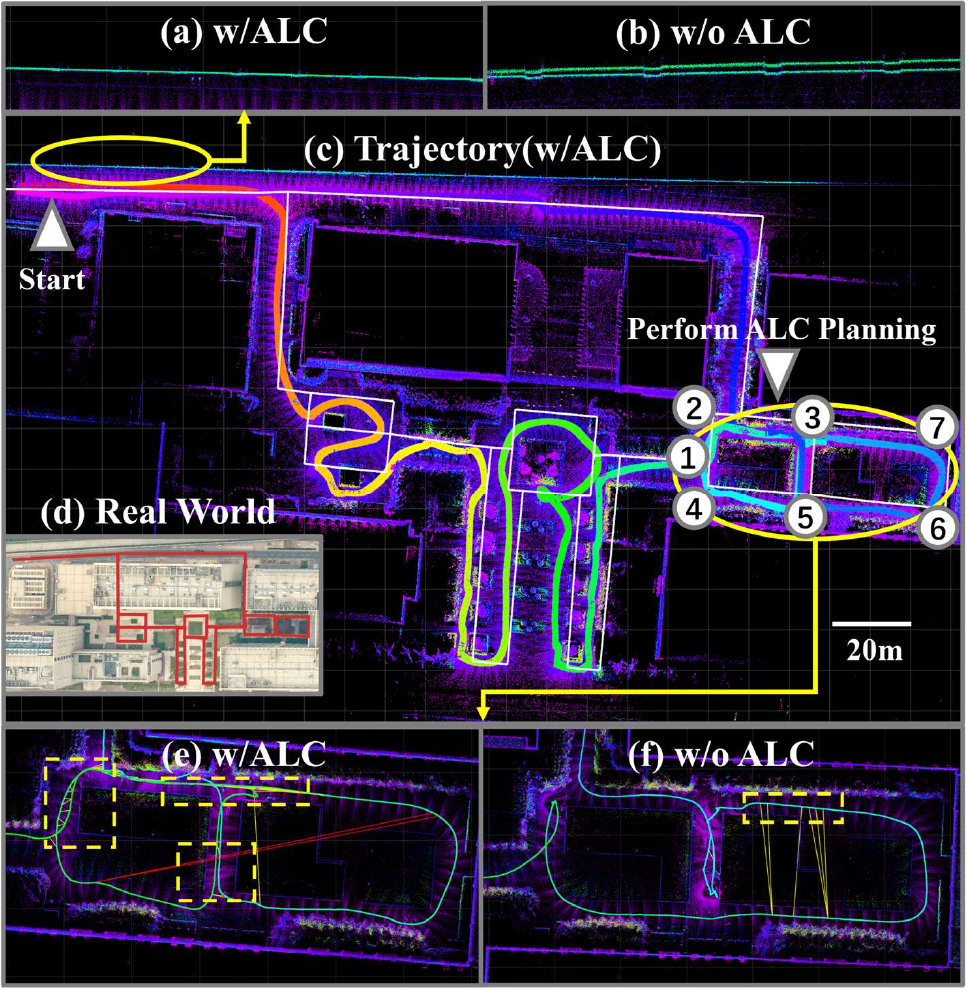}
    \caption{
    Experiment comparison results for Env.5. (a), (b): w/o ALC the map is distorted in the wall area. (c) The exploration trajectory of w/ALC, where the color of the trajectory indicates the order of exploration, and the white lines represent the prior OSM graph. (d) real world satellite images. (e), (f): w/ALC explored this area with routes: 1\(\rightarrow\)2\(\rightarrow\)3\(\rightarrow\)5\(\rightarrow\)4\(\rightarrow\)1\(\rightarrow\)2\(\rightarrow\)3\(\rightarrow\)7\(\rightarrow\)6\(\rightarrow\)5\(\rightarrow\)3\(\rightarrow\)start. w/o ALC explored this area with routes: 1\(\rightarrow\)2\(\rightarrow\)1\(\rightarrow\)4\(\rightarrow\)5\(\rightarrow\)6\(\rightarrow\)7\(\rightarrow\)3\(\rightarrow\)5\(\rightarrow\)3\(\rightarrow\)start. The yellow straight line in the yellow dashed box represents the occurrence of loop closure detection.}
    \label{experiment results}
    \vspace{-0.5cm}
\end{figure}

To evaluate the performance of our method, we deployed experiments in five representative environments. These environments include 1. Special terrain environments such as bridges, plazas, and extensive lawns, which is challenging for pose estimation (scene 1 and scene 2).
2. Environment with large OSM bias (scene 3). 
3. Environments with complex topology road networks (scene 4 and scene 5).
Before the mapping starts, we need to obtain the optimal global path from the global planning module, as described in Sec.\ref{sec: Global Candidate Path Generation}. It is noticeable that the number of solutions to the CPP increases rapidly with the increased complexity of the graph. It can take a long time to traverse a complex graph with a large number of nodes before obtaining the optimal solution. 
Practically, we employ random exhaustive search to iteratively find possible solutions, continuing this process until the number of solutions converges, thereby approximating the optimal route as defined by Eq.\ref{global eq}. Table \ref{table1} records the maximum and minimum $D\text{-}opt(\mathbf{L}_\gamma(\mathcal{R}))$  in different environments, which represent the best and worst routes, as well as the number of nodes, exhaustive attempts, and solutions. The maximum and minimum values for environment 2 correspond to routes (a) and (b) in Fig.\ref{eog}.
A comparison of mapping with the ALC module (w/ALC) to mapping based only on global planning routes (w/o ALC) for the global performance and local details of Env.5 is shown in Fig.\ref{experiment results}. Though the prior OSM graph is
complex, the robot still completes the
mapping task autonomously and
efficiently. The w/ALC approach results in
higher quality maps (Fig.\ref{experiment results} (a) v.s. (b)), as well as more efficient active loop closure routes planned in real time (Fig.\ref{experiment results} (g) v.s. (h)).

\begin{table}[ht]
\centering
\caption{Candidate Routes Comparison}
\label{table1}
\begin{tabular}{@{}cccccccc@{}}
\toprule
Env. & Max        & Min      & \#Nodes & \#Attempts & \#Solutions & Time (s)\\ 
\midrule
1    & 2.072 & 1.057 & 56      & 500        &  44      & 0.921 \\
2   & 4.272 & 2.250 & 58      & 500        &  60      & 0.962        \\
3    & 1.963     & 1.299     & 91       & 800     & 68       & 2.279   \\
4    & 5.408 & 4.057     & 134     & 8000   &416         & 31.566     \\
5    & 4.769    & 2.556     & 67    & 100000   & 5899      & 225.810      \\
\bottomrule
\end{tabular}
\end{table}

\begin{table*}[ht]
\centering
\caption{Quantitative results on uncertainty in robot poses and pose error reduction.}
\label{table analysis}
\small
\setlength{\tabcolsep}{10pt}
\begin{tabular}{@{}lccccccccc@{}}
\toprule
& \multicolumn{3}{c}{Uncertainty w/o ALC} & \multicolumn{3}{c}{Uncertainty w/ ALC} & \multicolumn{3}{c}{Pose Error Reduction (Trans. (m))} \\
\cmidrule(lr){2-4} \cmidrule(lr){5-7} \cmidrule(lr){8-10}
Env. & Avg. & Max & Traj.Len(m) & Avg. & Max & Traj.Len(m) & w/o ALC & w/ALC & Reduction \\
\midrule
1 & 0.131 & 0.340 & 1695 & 0.103 & 0.227 & 1647 & 4.429 & 1.615 & 63.54\% \\
2 & 0.172 & 0.351 & 1555 & 0.163 & 0.355 & 1567 & 2.764 & 1.649 & 40.33\% \\
3 & 0.141 & 0.298 & 1260 & 0.114 & 0.239 & 1245 & 1.353 & 0.621 & 54.08\% \\
4 & 0.105 & 0.374 & 1613 & 0.098 & 0.326 & 1609 & 1.543 & 1.022 & 33.79\% \\
5 & 0.096 & 0.211 & 895  & 0.088 & 0.177 & 995  & 1.408 & 0.865 & 38.60\% \\
\bottomrule
\end{tabular}
\end{table*}

\subsection{Evaluation of Active Loop Closure Module}

\begin{figure}[ht]
    \vspace{-0.2cm}
    \centering
    \vspace{-0.1cm}
    \subfigure[Env. 1]{
        \includegraphics[width=0.46\linewidth,trim=7 0 6 0,clip]{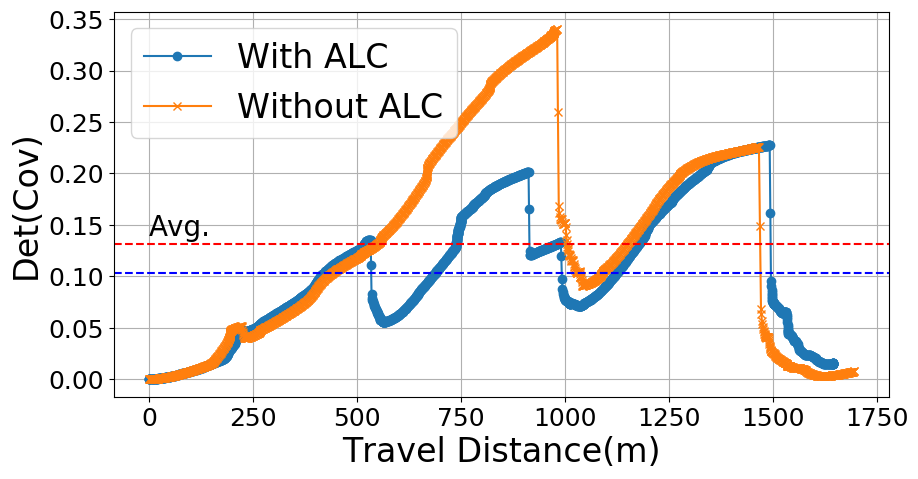}
    }
    \vspace{-0.1cm}
    \subfigure[Env. 2]{
        \includegraphics[width=0.46\linewidth,trim=7 0 6 0,clip]{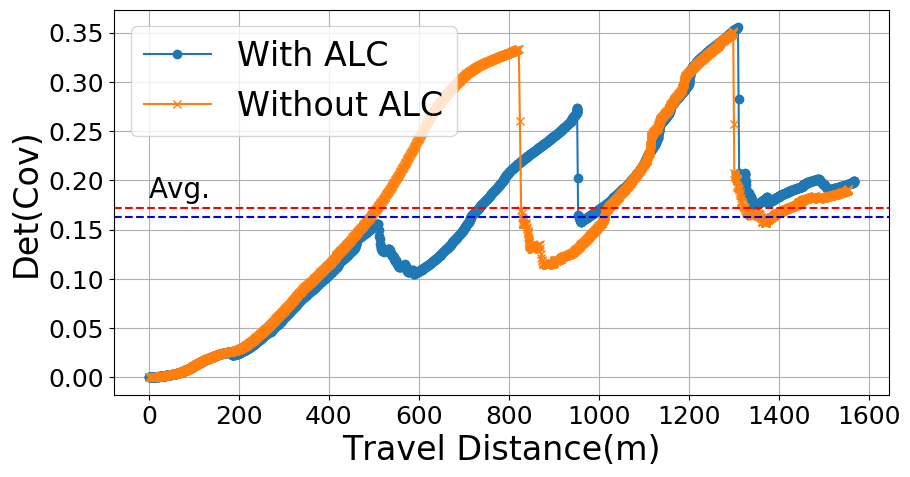}
    }
    \vspace{-0.1cm}
    \subfigure[Env. 3]{
        \includegraphics[width=0.46\linewidth,trim=7 0 6 0,clip]{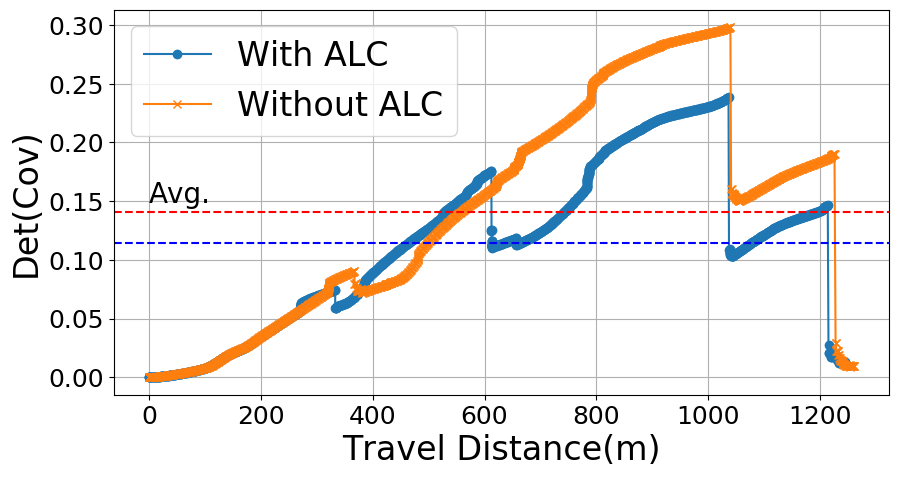}
    }
    \vspace{-0.1cm}
    \subfigure[Env. 4]{
        \includegraphics[width=0.46\linewidth,trim=7 0 6 0,clip]{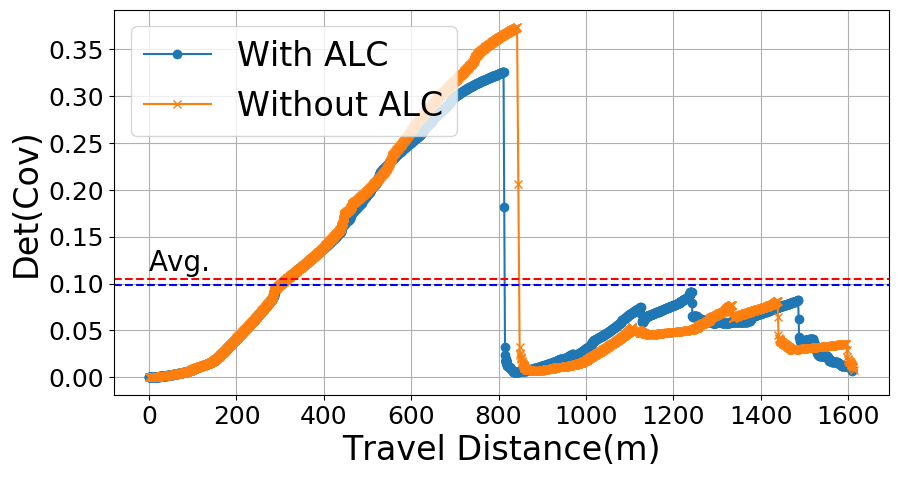}
    }
    \vspace{-0.1cm}
    \subfigure[Env. 5]{
        \includegraphics[width=0.46\linewidth,trim=7 0 6 0,clip]{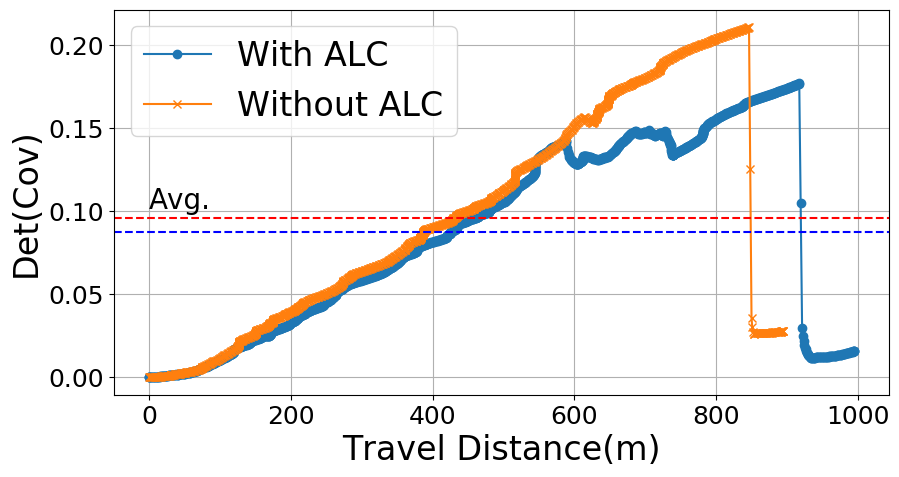}
    }
    \caption{Instant uncertainty with travel distance
    }
    \label{uncertainty output}
    \vspace{-0.5cm}
\end{figure}

We compared the pose estimation error and uncertainty with and without the ALC module under the same parameter settings to evaluate its effectiveness. Firstly, we performed a quantitative analysis comparing the variation of uncertainty with respect to the travel distance during the mapping process in each scenario, where the uncertainty was calculated from the determinant of the marginal covariance matrix associated with the current pose, as shown in Fig.\ref{uncertainty output}. The rapid decrease in the plot represents an occurrence of loop closure. It can be seen that compared to the w/o ALC approach, the w/ALC approach can find the loop closure much faster and keep the average and maximum uncertainty of the pose in a low range while maintaining an approximate equal-length trajectory (shown in Tab.\ref{table analysis}). In environments 4 and 5, due to the complexity of the topology, the trajectories provided by global planning have multiple duplicated edges or nodes for loop closure detection, resulting in lower average uncertainty.

We utilized a GPS RTK to provide ground truth and quantitatively compare the
instant absolute pose error (APE) by
using \textit{evo}\footnote{https://github.com/MichaelGrupp/evo}. Note that we evaluated the online position accuracy, not the keyframe-pose accuracy after the complete loop closure optimization, which is meaningful for analyzing the pose drift of the robot during the online mapping. As shown in Tab.\ref{table analysis}, our ALC module effectively reduces the robot's translation error by 33.79\%-63.54\%. In our experimental measurements, we found that the maximum bias in position estimation in special terrain environment can be so high (Env.1 8.87m, Env.2 5.28m) as to make it difficult for the robot to navigate safely and accurately, while the addition of the ALC module reduces this maximum position error to an acceptable value(Env.1 3.31m, Env.2 3.07m).
These quantitative results show that our method enables robots to construct high-quality maps efficiently and accurately.

\section{Conclusion}
This work exploits OSM prior information to achieve real-time active SLAM for robots in real-world large-scale urban scenes. Our approach integrates active loop closure strategies with global path planning based on OSM data, significantly reducing the uncertainty in pose estimation that is common in large-scale SLAM scenarios. The implementation on a quadruped robot and extensive field tests in diverse urban environments demonstrate the system's effectiveness in navigating and mapping with precision.

% \bibliographystyle{unsrt}
% \bibliography{irosRefs}

\begin{thebibliography}{10}

\bibitem{yamauchi1997frontier}
Brian Yamauchi.
\newblock A frontier-based approach for autonomous exploration.
\newblock In {\em Proceedings 1997 IEEE International Symposium on Computational Intelligence in Robotics and Automation CIRA'97.'Towards New Computational Principles for Robotics and Automation'}, pages 146--151, 1997.

\bibitem{keidar2014efficient}
Matan Keidar and Gal~A Kaminka.
\newblock Efficient frontier detection for robot exploration.
\newblock {\em The International Journal of Robotics Research}, pages 215--236, 2014.

\bibitem{cao2021tare}
Chao Cao, Hongbiao Zhu, Howie Choset, and Ji~Zhang.
\newblock Tare: A hierarchical framework for efficiently exploring complex 3d environments.
\newblock In {\em Robotics: Science and Systems}, 2021.

\bibitem{kulkarni2022autonomous}
Mihir Kulkarni, Mihir Dharmadhikari, Marco Tranzatto, Samuel Zimmermann, Victor Reijgwart, Paolo De~Petris, Huan Nguyen, Nikhil Khedekar, Christos Papachristos, Lionel Ott, et~al.
\newblock Autonomous teamed exploration of subterranean environments using legged and aerial robots.
\newblock In {\em 2022 International Conference on Robotics and Automation (ICRA)}, pages 3306--3313, 2022.

\bibitem{sun2020frontier}
Zezhou Sun, Banghe Wu, Cheng-Zhong Xu, Sanjay~E Sarma, Jian Yang, and Hui Kong.
\newblock Frontier detection and reachability analysis for efficient 2d graph-slam based active exploration.
\newblock In {\em 2020 IEEE/RSJ Intl. Conf. on Intell. Robots and Syst. (IROS)}, pages 2051--2058, 2020.

\bibitem{zhu2021dsvp}
Hongbiao Zhu, Chao Cao, Yukun Xia, Sebastian Scherer, Ji~Zhang, and Weidong Wang.
\newblock Dsvp: Dual-stage viewpoint planner for rapid exploration by dynamic expansion.
\newblock In {\em 2021 IEEE/RSJ Intl. Conf. on Intell. Robots and Syst. (IROS)}, pages 7623--7630, 2021.

\bibitem{sun2023concave}
Zezhou Sun, Banghe Wu, Chengzhong Xu, and Hui Kong.
\newblock Concave-hull induced graph-gain for fast and robust robotic exploration.
\newblock {\em IEEE Robotics and Automation Letters}, 2023.

\bibitem{soragna2019active}
Alberto Soragna, Marco Baldini, Dominik Joho, Rainer K{\"u}mmerle, and Giorgio Grisetti.
\newblock Active slam using connectivity graphs as priors.
\newblock In {\em 2019 IEEE/RSJ Intl. Conf. on Intell. Robots and Syst. (IROS)}, pages 340--346, 2019.

\bibitem{xu2021crmi}
Yang Xu, Ronghao Zheng, Meiqin Liu, and Senlin Zhang.
\newblock Crmi: Confidence-rich mutual information for information-theoretic mapping.
\newblock {\em IEEE Robotics and Automation Letters}, pages 6434--6441, 2021.

\bibitem{bai2023graph}
Ruofei Bai, Hongliang Guo, Wei-Yun Yau, and Lihua Xie.
\newblock Graph-based slam-aware exploration with prior topo-metric information.
\newblock {\em arXiv preprint arXiv:2308.16522}, 2023.

\bibitem{jungnickel2005graphs}
Dieter Jungnickel and D~Jungnickel.
\newblock {\em Graphs, networks and algorithms}.
\newblock Springer, 2005.

\bibitem{umari2017autonomous}
Hassan Umari and Shayok Mukhopadhyay.
\newblock Autonomous robotic exploration based on multiple rapidly-exploring randomized trees.
\newblock In {\em 2017 IEEE/RSJ Intl. Conf. on Intell. Robots and Syst. (IROS)}, pages 1396--1402, 2017.

\bibitem{placed2023survey}
Julio~A Placed, Jared Strader, Henry Carrillo, Nikolay Atanasov, Vadim Indelman, Luca Carlone, and Jos{\'e}~A Castellanos.
\newblock A survey on active simultaneous localization and mapping: State of the art and new frontiers.
\newblock {\em IEEE Transactions on Robotics}, 2023.

\bibitem{stachniss2005information}
Cyrill Stachniss, Giorgio Grisetti, and Wolfram Burgard.
\newblock Information gain-based exploration using rao-blackwellized particle filters.
\newblock In {\em Robotics: Science and systems}, pages 65--72, 2005.

\bibitem{zhang2022exploration}
Yichen Zhang, Boyu Zhou, Luqi Wang, and Shaojie Shen.
\newblock Exploration with global consistency using real-time re-integration and active loop closure.
\newblock In {\em 2022 International Conference on Robotics and Automation (ICRA)}, pages 9682--9688, 2022.

\bibitem{tao20243d}
Yuezhan Tao, Xu~Liu, Igor Spasojevic, Saurav Agarwal, and Vijay Kumar.
\newblock 3d active metric-semantic slam.
\newblock {\em IEEE Robotics and Automation Letters}, 2024.

\bibitem{osswald2016speeding}
Stefan O{\ss}wald, Maren Bennewitz, Wolfram Burgard, and Cyrill Stachniss.
\newblock Speeding-up robot exploration by exploiting background information.
\newblock {\em IEEE Robotics and Automation Letters}, pages 716--723, 2016.

\bibitem{eiselt1995arc}
Horst~A Eiselt, Michel Gendreau, and Gilbert Laporte.
\newblock Arc routing problems, part i: The chinese postman problem.
\newblock {\em Operations Research}, pages 231--242, 1995.

\bibitem{placed2021fast}
Julio~A Placed and Jos{\'e}~A Castellanos.
\newblock Fast autonomous robotic exploration using the underlying graph structure.
\newblock In {\em 2021 IEEE/RSJ Intl. Conf. on Intell. Robots and Syst. (IROS)}, pages 6672--6679, 2021.

\bibitem{carrillo2012comparison}
Henry Carrillo, Ian Reid, and Jos{\'e}~A Castellanos.
\newblock On the comparison of uncertainty criteria for active slam.
\newblock In {\em 2012 IEEE International Conference on Robotics and Automation}, pages 2080--2087, 2012.

\bibitem{pazman1986foundations}
Andrej P{\'a}zman.
\newblock Foundations of optimum experimental design.
\newblock {\em (No Title)}, 1986.

\bibitem{khosoussi2019reliable}
Kasra Khosoussi, Matthew Giamou, Gaurav~S Sukhatme, Shoudong Huang, Gamini Dissanayake, and Jonathan~P How.
\newblock Reliable graphs for slam.
\newblock {\em The International Journal of Robotics Research}, pages 260--298, 2019.

\bibitem{grisetti2010tutorial}
Giorgio Grisetti, Rainer K{\"u}mmerle, Cyrill Stachniss, and Wolfram Burgard.
\newblock A tutorial on graph-based slam.
\newblock {\em IEEE Intelligent Transportation Systems Magazine}, pages 31--43, 2010.

\bibitem{sorenson1980parameter}
Harold~Wayne Sorenson.
\newblock Parameter estimation: principles and problems.
\newblock {\em (No Title)}, 1980.

\bibitem{placed2022general}
Julio~A Placed and Jos{\'e}~A Castellanos.
\newblock A general relationship between optimality criteria and connectivity indices for active graph-slam.
\newblock {\em IEEE Robotics and Automation Letters}, pages 816--823, 2022.

\bibitem{xu2022fast}
Wei Xu, Yixi Cai, Dongjiao He, Jiarong Lin, and Fu~Zhang.
\newblock Fast-lio2: Fast direct lidar-inertial odometry.
\newblock {\em IEEE Transactions on Robotics}, pages 2053--2073, 2022.

\bibitem{wang2020lidar}
Ying Wang, Zezhou Sun, Cheng-Zhong Xu, Sanjay~E Sarma, Jian Yang, and Hui Kong.
\newblock Lidar iris for loop-closure detection.
\newblock In {\em 2020 IEEE/RSJ Intl. Conf. on Intell. Robots and Syst. (IROS)}, pages 5769--5775, 2020.

\bibitem{dellaert2012factor}
Frank Dellaert.
\newblock Factor graphs and gtsam: A hands-on introduction.
\newblock {\em Georgia Institute of Technology, Tech. Rep}, page~4, 2012.

\bibitem{cao2022autonomous}
Chao Cao, Hongbiao Zhu, Fan Yang, Yukun Xia, Howie Choset, Jean Oh, and Ji~Zhang.
\newblock Autonomous exploration development environment and the planning algorithms.
\newblock In {\em 2022 International Conference on Robotics and Automation (ICRA)}, pages 8921--8928, 2022.

\end{thebibliography}

\end{document}